\documentclass[journal]{IEEEtran}

\usepackage{cite}
\usepackage{amsmath,amssymb,amsfonts}
\usepackage{algorithmic}
\usepackage{graphicx}
\usepackage{textcomp}
\usepackage{bm}
\usepackage{makecell}
\usepackage[normalem]{ulem}
\usepackage{multirow}
\usepackage{xcolor}
\usepackage{tabularx,threeparttable}
\usepackage[ruled,linesnumbered]{algorithm2e}
\usepackage{booktabs}
\usepackage{arydshln}
\usepackage{comment}

\begin{document}
\title{Instrument-Splatting++: Towards Controllable Surgical Instrument Digital Twin Using Gaussian Splatting}
\author{Shuojue Yang$^\ast$, Zijian Wu$^\ast$,\IEEEmembership{ Graduate Student Member, IEEE}, Chengjiaao Liao, Qian Li, Daiyun Shen, \\ Chang Han Low, Septimiu E. Salcudean, \IEEEmembership{ Life Fellow, IEEE} and Yueming Jin$^{\ast\ast}$, \IEEEmembership{Member, IEEE}
\thanks{S. Yang, C. Liao, D. Shen, Q. Li, C. Low and Y. Jin are with Department of Biomedical Engineering, National University of Singapore (NUS), Singapore. Y. Jin is also with Department of Electrical and Computer Engineering, NUS. (email:\{s.yang, e1454387, e1374467,e0543455\}@nus.edu.sg, \{liqian, ymjin\}@nus.edu.sg)}
\thanks{Z. Wu, S. Salcudean are with Department of Electrical and Computer Engineering, The University of British Columbia, Vancouver, Canada (email: \{zijianwu, tims\}@ece.ubc.ca)}
\thanks{* Shuojue Yang and Zijian Wu are equal contributors.}
\thanks{** Corresponding author: Yueming Jin. (e-mail: ymjin@nus.edu.sg)}
}

\maketitle

\begin{abstract}
High-quality and controllable digital twins of surgical instruments are critical for Real2Sim in robot-assisted surgery, as they enable realistic simulation, synthetic data generation, and perception learning under novel poses. We present \textit{Instrument-Splatting++}, a monocular 3D Gaussian Splatting (3DGS) framework that reconstructs surgical instruments as a fully controllable Gaussian asset with high fidelity. Our pipeline starts with part-wise geometry pretraining that injects CAD priors into Gaussian primitives and equips the representation with part-aware semantic rendering. Built on the pretrained model, we propose a semantics-aware pose estimation and tracking (SAPET) method to recover per-frame 6-DoF pose and joint angles from unposed endoscopic videos, where a gripper-tip network trained purely from synthetic semantics provides robust supervision and a loose regularization suppresses singular articulations. Finally, we introduce Robust Texture Learning (RTL), which alternates pose refinement and robust appearance optimization, mitigating pose noise during texture learning. The proposed framework can perform pose estimation and learn realistic texture from unposed videos. We validate our method on sequences extracted from EndoVis17/18, SAR-RARP, and an in-house dataset, showing superior photometric quality and improved geometric accuracy over state-of-the-art baselines. We further demonstrate a downstream keypoint detection task where unseen-pose data augmentation from our controllable instrument Gaussian improves performance.
\end{abstract}


\section{Introduction}
\label{sec:introduction}
\IEEEPARstart{R}{obot}-assisted surgery has gained increasing adoption due to improved dexterity, reduced intraoperative trauma, and better clinical outcomes~\cite{wah2025rise}. Modern systems such as the da Vinci platform provide high-resolution 3D visualization and precise teleoperation, motivating growing interest in AI-enabled autonomous or semi-autonomous surgical assistance~\cite{iftikhar2024artificial}. Central to this vision is a high-quality digital twin of articulated instruments, enabling embodied surgical manipulation. Ideally, such digital twins should achieve accurate appearance and geometry, as well as high controllability, which allows reposing the instrument digital twin to arbitrary configurations via user-defined kinematic parameters, and thus enables interactive simulation and supports sim-to-real policy learning~\cite{ou2023sim,scheikl2022sim} and synthetic data generation for surgical perception~\cite{cartucho2021visionblender,ding2024digital}. However, real endoscopic videos exhibit rapid camera and instrument motion as well as complex illumination, which are inadequately captured by existing simulators, resulting in a huge domain gap between simulated and real surgical data.

To address this issue, there have been great efforts in Real2Sim~\cite{wang2025embodiedreamer,lou2025robo} that learns controllable robots and environments from real video and inject them into simulators. This line of work targets a minimized sim-to-real gap, enhancing both robotic manipulation and perception. 
Particularly, 3D Gaussian Splatting (GS)~\cite{kerbl20233d} has shown as a controllable robot representation in several general robotics work~\cite{liu2024differentiable,lou2025robo,lu2024manigaussian}, which provides a high visual fidelity while preserving controllability that allows users to interactively manipulate the simulated robots. 
\begin{figure}[t]
	\centering
	\includegraphics[width=0.9\columnwidth]{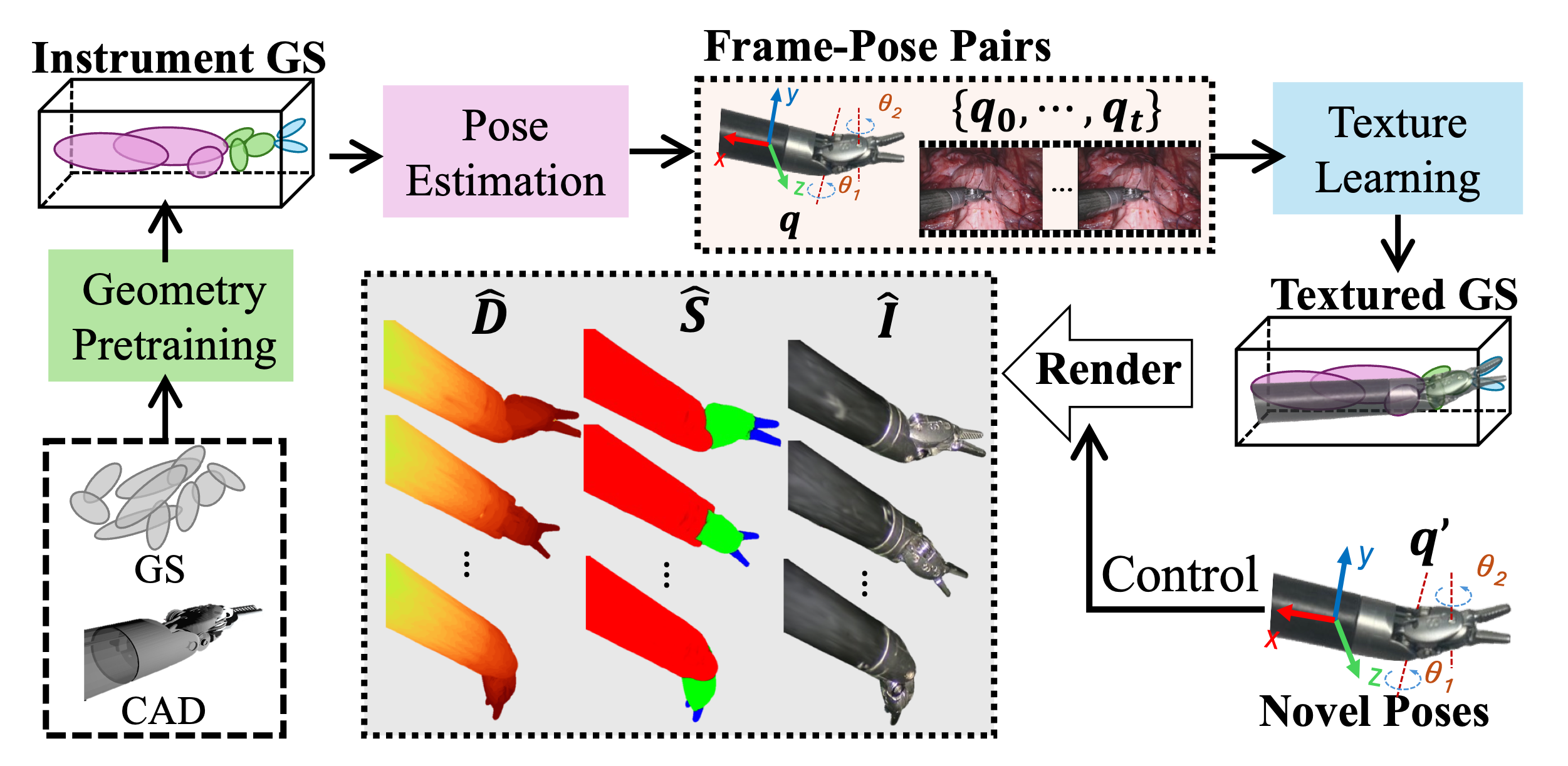}
	\vspace{-3mm}
	\caption{Overall pipeline of our Real2Sim pipeline \textit{Instrument-Splatting++}, which yields controllable GS digital twin of surgical instrument (Textured GS) with high visual and geometric fidelity.}
	\label{fig:teaser}
	\vspace{-3mm}
\end{figure}
Nevertheless, GS-based 3D reconstruction methods~\cite{yang2024deform3dgs,liu2024endogaussian,chen2025surgicalgs,xie2024surgicalgaussian} in surgical domain often exclude instruments and focus primarily on reconstructing the deformable tissues. Besides, deformation fields in these dynamic GS methods are conditioned solely on timestamps, and therefore can only replay dynamic scenes rather than providing controllable 3D GS for surgical instruments.

It is also challenging to directly leverage GS-based methods for general robot or articulated object reconstruction in surgical domain. 
First, most prior work~\cite{liu2025building,guo2025articulatedgs,kim2025screwsplat,lin2025splart} heavily relies on a static initialization stage where the articulated robot will remain static and undergo multi-view scans, which are not practical for surgical setting with a single viewpoint. 
Also, learning 3D GS-based controllable robot representations~\cite{lou2025robo,zeng2025neeco,liu2024differentiable} typically requires images paired with robot articulated pose annotations to reach high reconstruction performance while maintaining controllability,however, due to kinematics errors and patient privacy for surgical robots, acquisition of accurate pose data is largely limited~\cite{allan20183,cui2023caveats}. 

Driven by these challenges, we present \textit{Instrument-Splatting++}, a novel 3D GS-based Real2Sim framework for articulated and controllable reconstruction of surgical instruments. As shown in Fig.~\ref{fig:teaser}, our pipeline begins with geometry pretraining that injects accurate geometric priors from CAD models into Gaussians for each rigid part, which are then assembled into a unified instrument GS. This initialization ensures geometric accuracy and eliminates the need for complex multi-view static initialization from video.
Building upon the pretrained Gaussians, we propose a Semantics-Aware Pose Estimation and Tracking (SAPET) strategy to recover the per-frame instrument poses from monocular videos in a render-and-compare manner which minimizes the discrepancy between rendered part-level semantic maps and segmentation masks. Moreover, we additionally introduce a Gripper-Tip Net trained on synthetic data to reliably estimate 2D tip points from segmentation masks only, which serves to guide the pose optimization together with a structural prior-based loose regularization. Subsequently, we propose a novel Robust Texture Learning (RTL) strategy to mutually reinforce the pose rectification and texture learning, where poses and GS appearances are alternately optimized to reach photorealism of reconstruction even under pose noises. Extensive experiments show that \textit{Instrument-Splatting++} can recover accurate articulated poses and photorealistic textures from unposed monocular videos, while also enabling real-time rendering and benefiting downstream keypoint detection through unseen-pose data augmentation. This proposed Real2Sim pipeline demonstrates large potential in benefiting surgical Artificial Intelligence (AI) and autonomy by providing video-kinematics paired data for robot imitation learning~\cite{chen2025surgipose} and high-fidelity synthetic data augmentation~\cite{zeng2024realistic}.

This work substantially extends our preliminary work (Yang et al.~\cite{yang2025instrument}) on surgical instrument reconstruction at MICCAI 2025. First, during pose estimation stage, different from previous Singular Value Decomposition (SVD) based gripper tip detection method, we introduce a  Gripper-Tip Net that takes part-level segmentation masks as input, and directly predicts the gripper tip locations, significantly improving the detection reliability. Besides, our previous work solely uses silhouette maps to supervise the pose optimization, ignoring the learned textures with rich visual cues. Also, the previous work directly uses estimated poses to perform the texture learning, which overlooks potential noise in estimated poses and thus leads to degraded reconstruction performance. In this work, we introduce a novel Robust Texture Learning (RTL) strategy,  in which a pose refinement step leverages learned textures to progressively correct noisy estimated poses, and a robust Gaussian appearance learning scheme alleviates the impact of pose noises during texture learning. 
Furthermore, we conduct extensive experiments on one more intraoperative dataset (SAR-RARP) for comprehensive evaluation. We newly add rendered depth comparison to validate the 3D geometry fidelity of the proposed method. 
We further validate the effectiveness of our method in supporting a crucial downstream task of keypoint detection. Our method enables unseen-pose data synthesis, which facilitates more accurate keypoint detection. The code and dataset for reconstructing the instrument digital twin will be released. Also, the image-pose paired dataset included in this work will be released for potential robot learning applications.

\section{Related Works}
Surgical scene reconstruction has increasingly adopted differentiable rendering techniques for their ability to produce photorealistic results. Early NeRF-based methods~\cite{mildenhall2021nerf} demonstrated strong visual quality but were limited by training efficiency~\cite{wang2022neural}. More recently, 3D Gaussian Splatting (3DGS)~\cite{kerbl20233d} has emerged as a dominant representation, enabling faster training and higher-fidelity reconstruction. While these approaches improve geometric accuracy~\cite{chen2025surgicalgs}, lighting robustness~\cite{kaleta2025pr}, or deformation modeling~\cite{xu2025t}, they primarily focus on soft-tissue reconstruction and largely ignore surgical instruments. Due to the fast, articulated, and dexterous motions of instruments, existing deformation-field formulations struggle to capture their dynamics. Moreover, prior work targets high-fidelity scene replay rather than producing controllable digital twins suitable for interactive simulation.

Recently, several studies have begun to explore surgical instrument reconstruction. Zeng \emph{et al.}~\cite{zeng2024realistic} employ 3D GS to reconstruct instruments from multiple viewpoints and use the resulting models for data augmentation in downstream segmentation tasks; however, the instrument is treated as a rigid object, which significantly limits its controllability. Similarly, Li \emph{et al.}~\cite{li20243dgs} perform a static GS reconstruction stage to jointly recover instrument geometry, appearance, and articulated part information, and subsequently apply differentiable rendering for pose estimation. However, their reconstruction relies on multi-view capture of static instruments, which limits its applicability to real clinical surgical videos. NeeCo~\cite{zeng2025neeco} introduces a deformation field to model articulation changes of hand-held surgical instruments, yet relies on external electromagnetic tracking to obtain accurate instrument and endoscope poses, restricting its applicability to recorded clinical videos. More recently, Xu \emph{et al.}~\cite{xu2025t} leverage point tracks to globally supervise fast instrument motion in dynamic scene reconstruction, but do not explicitly model articulated instrument kinematics and cannot provide controllable 3D digital assets.



\begin{figure*}[t]
    \centering
    \includegraphics[width=0.9\textwidth]{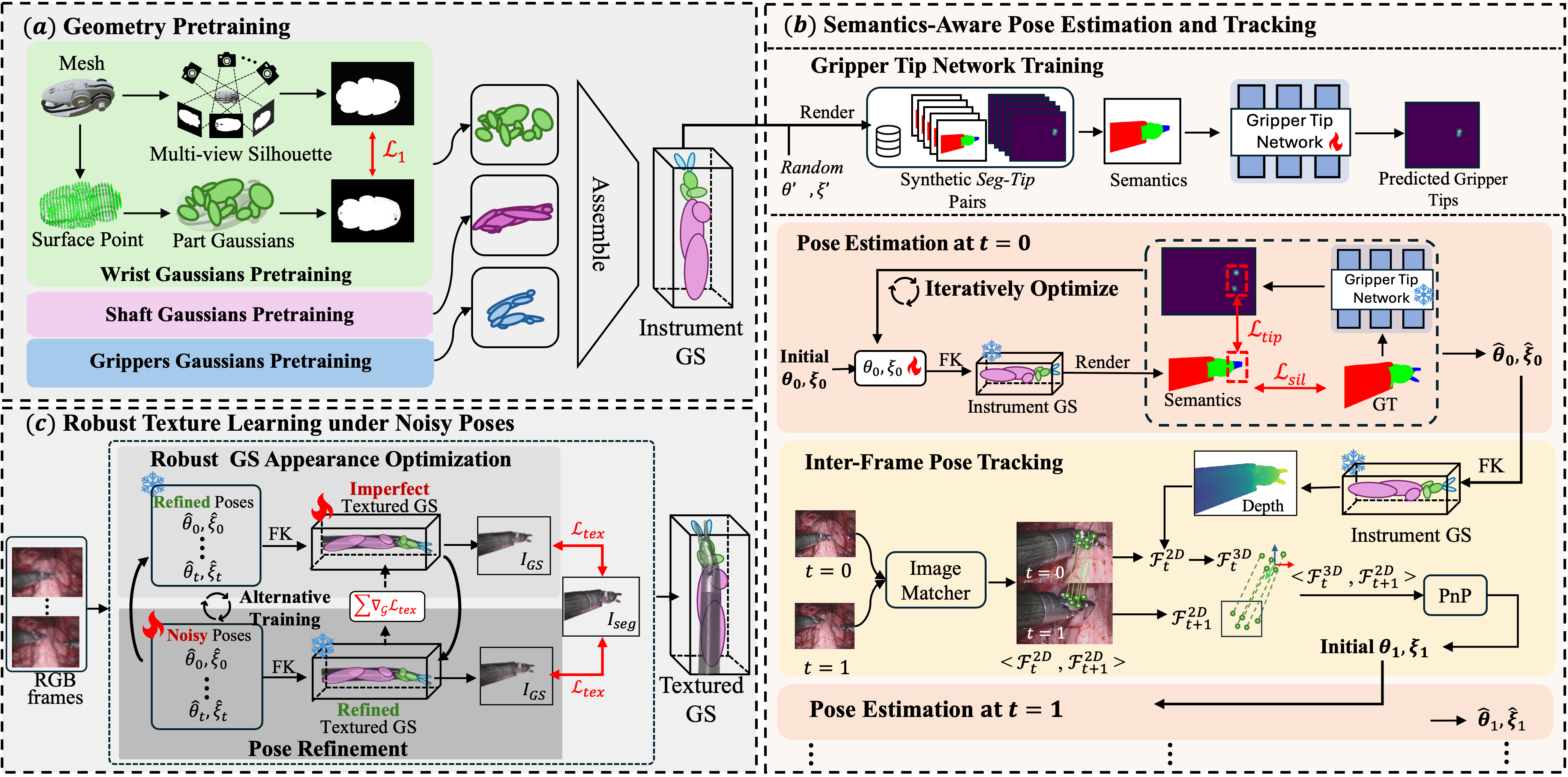} 
    \caption{Overview of the proposed Instrument-Splatting++. Black arrows indicate operation flow, while red arrows denote supervisory signals. The textured GS in \textit{(c)} denotes the generated GS digital twin of instrument with learned textures. Fire and snowflake icons represent optimizable and frozen parameters, respectively.}
    \label{fig:method}
\end{figure*}

\section{Methods}
The overview of our method is illustrated in Fig. \ref{fig:method}. 
In this section, we first briefly describe the details of the proposed method. Our approach begins with a geometry pretraining to initialize the Gaussian points with accurate geometric prior (Sec. \ref{sec:3.2}). 
Subsequently, we propose a frame-by-frame pose estimation and tracking method (SAPET) to estimate per-frame instrument poses, enhanced by a gripper tip detection network trained with only synthetic dataset (Sec. \ref{sec:3.3}).
Lastly, we develop a robust texture learning strategy to learn GS appearance under noisy estimated poses (Sec. \ref{sec:3.4}).
\begin{figure}[t]
	\centering
	\includegraphics[width=0.9\columnwidth]{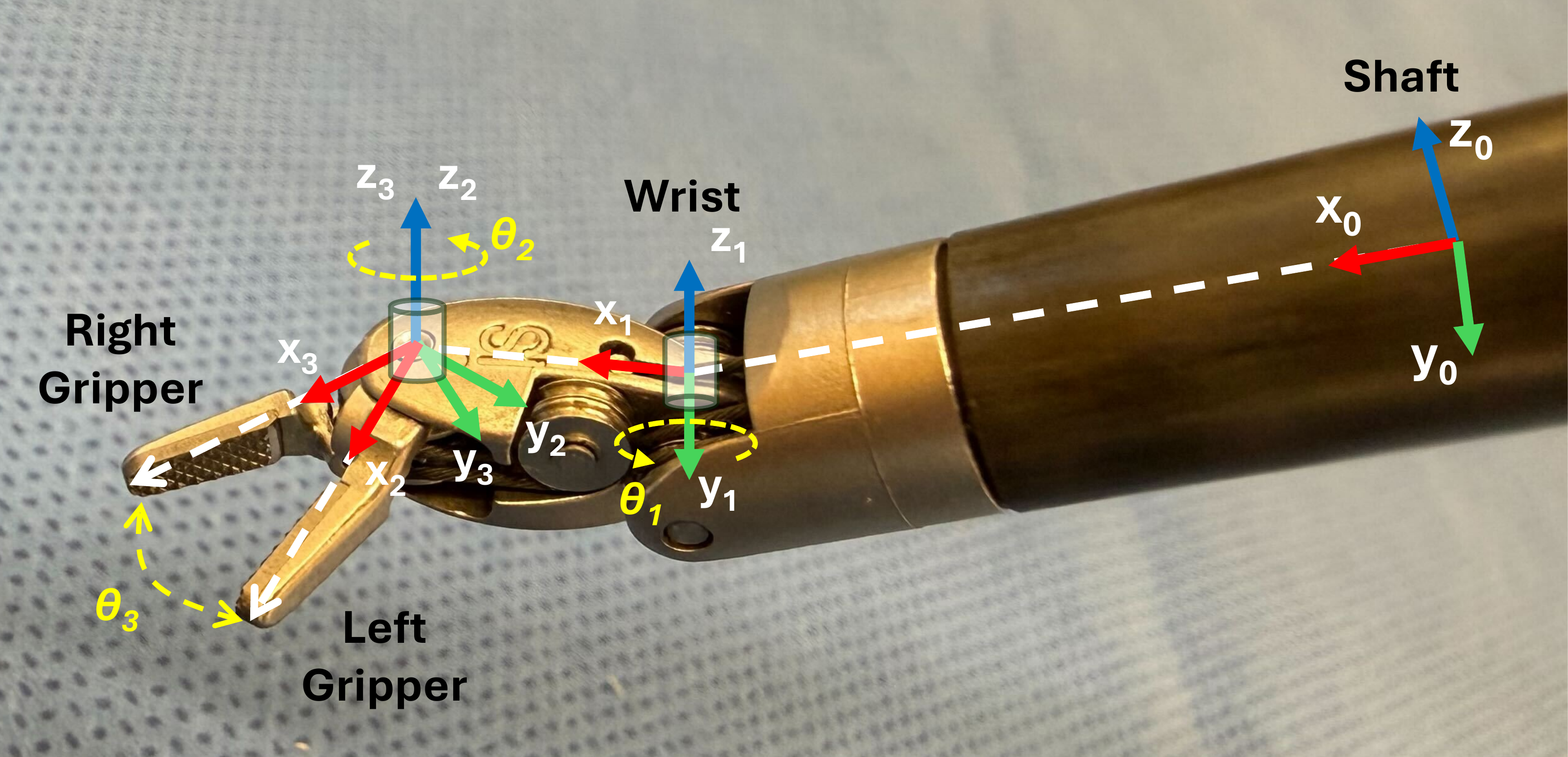}
	\vspace{-3mm}
	\caption{Forward kinematics definition of the da Vinci EndoWrist Large Needle Driver (LND).}
	\label{fig:kinematics}
	\vspace{-3mm}
\end{figure}
\subsection{Geometry Pretraining}
\label{sec:3.2}
Our method starts with Geometry Pretraining where a geometrically accurate part Gaussian model will be initialized from the CAD models. With a well-designed kinematics, we can assemble these rigid part Gaussians into a controllable articulated instrument Gaussian model.\\
\textbf{Kinematics Design.}
We model articulated surgical instruments as a kinematic chain composed of rigid links connected by revolute joints, following the mechanical design of surgical tools. In this work, the da Vinci EndoWrist Large Needle Driver (LND) is used as a representative example, for which an open-source CAD model is available. 

As illustrated in Fig.~\ref{fig:kinematics}, the instrument is described by a set of local coordinate frames attached to its main functional parts. The shaft frame is treated as the base frame, while subsequent frames correspond to the wrist, right gripper, and left tooltip. We unify both articulation and global pose into a single pose set $\mathbf{q} = \{\theta_1, \theta_2, \theta_3, \boldsymbol{\xi}\},$ where $\theta_1$, $\theta_2$, and $\theta_3$ denote the joint states of the LND and $\boldsymbol{\xi} \in \mathfrak{se}(3)$ parameterizes the rigid transformation from the shaft frame to the camera frame.

The forward kinematics of the articulated structure is defined through a product of rigid transformations. Given a part frame $j \in \{w, g, l\}$ (wrist, right gripper, and left tooltip), the transformation from the shaft frame to frame $j$ is written as
\begin{equation}
{}^{s}\mathbf{T}_j(\boldsymbol{\theta}) =
\begin{cases}
{}^{s}\mathbf{T}_w(\theta_1), 
& j = w \\[4pt]
{}^{s}\mathbf{T}_w(\theta_1)\,{}^{w}\mathbf{T}_g(\theta_2), 
& j = g \\[4pt]
{}^{s}\mathbf{T}_w(\theta_1)\,{}^{w}\mathbf{T}_g(\theta_2)\,{}^{g}\mathbf{T}_l(\theta_3), 
& j = l
\end{cases}
\label{eq1}
\end{equation}
The global pose of the instrument is captured by the rigid transformation ${}^{\text{cam}}\mathbf{T}_s(\boldsymbol{\xi})$.

Combining articulation and global motion,  a 3D point $\mathbf{x}_j\in\mathbb{R}^3$ defined in the local coordinate frame of part $j$ is mapped to the camera coordinate system by:
\begin{equation}
\mathbf{T}_j(\mathbf{x}_j;\mathbf{q}) 
= {}^{\text{cam}}\mathbf{T}_s(\boldsymbol{\xi}) \;
{}^{s}\mathbf{T}_j(\boldsymbol{\theta})\;
\mathbf{x}_j
\label{eq:fk_unified}
\end{equation}
This unified kinematic formulation updates the instrument GS primitives' positions $\boldsymbol{\mu}_j\in\mathbb{R}^3$ and $\mathbf{r}_j\in\mathbb{R}^4$ by $\mathbf{T}_j(\boldsymbol{\mu}_j;\mathbf{q})$ and $\mathbf{R}_j(\mathbf{r}_j;\mathbf{q})$, respectively, where $\mathbf{R}_j(\cdot)$ denotes the rotational component of $\mathbf{T}_j$.\\
\textbf{Geometry-Guided Initialization.}
To initialize the Gaussian models with accurate geometry, we perform geometry pretraining for each rigid part. 
For each rigid part, a set of Gaussian primitives is initialized in corresponding local coordinate frame. Given per-part CAD models, we initialize Gaussian points by uniformly sampling mesh surface points with ray-based sampling, which avoids sparse distributions caused by large mesh faces. To supervise the pretraining, binary silhouette images of each part are rendered from multiple viewpoints using Blender. During pretraining, all the GS parameters are set as optimizable, and an $\ell_1$ loss of the rendered opacity and ground-truth (GT) silhouette are minimized, encouraging geometrically accurate GS representation of each rigid part.

After pretraining, Gaussians from all rigid parts are aggregated into a unified instrument representation and  all the Gaussian primitives can be reposed using the kinematic chain defined in Eq.~\ref{eq:fk_unified}. Also, each Gaussian primitive is assigned with a semantic part label (i.e., shaft, wrist, or left/right gripper), allowing for differentiable part-aware semantics rendering $\widehat{\mathbf{S}}$ w.r.t. poses $\mathbf{q}$. 
%
\subsection{Semantics-Aware Pose Estimation and Tracking}
\label{sec:3.3}
\subsubsection{Gripper Tip Detection Network}
Gripper is the smallest part of surgical instruments and are dexterously moved during surgery. Due to its small size and rapid motion, locating grippers is technically challenging for iterative pose optimization that easily fails with no overlapped regions. Thus, gripper tips can serve as a global supervision to guide the instrument pose estimation. Different from Instrument-Splatting~\cite{yang2025instrument} that uses heuristic based SVD method to detect tip points, we introduce a Gripper-Tip Net that detects the gripper tip positions directly from segmentation masks to robustly guide the instrument pose across surgical sequences.\\
\textbf{Segmentation-based Simulated Data Generation.}
Leveraging the accurate part-aware semantic rendering of our controllable instrument Gaussians, we synthesize a large-scale dataset of 20,000 accurate semantics–tip pairs $(\mathbf{S}_i, \mathbf{p}_{i,\text{tip}})$ across diverse randomly generated instrument poses $\mathbf{q}$, where gripper tip locations $\mathbf{p}_{i,\text{tip}}$ are computed via the forward kinematic chain and projected onto the 2D image plane, and part-aware semantics maps $\mathbf{S}_i$ are rendered given part labels for each GS primitive and the geometry-pretrained Gaussian model. Of note, this synthetic dataset contains semantic maps only, thereby avoiding RGB images that will potentially introduce appearance bias during model training.\\
\textbf{Network Training.}
Our gripper tip network adopts a lightweight encoder–decoder architecture that regresses 2D gripper tip heatmaps from part-aware semantic inputs only. Following~\cite{wu2025tooltipnet}, we employ ResNet-50~\cite{he2016deep} as the encoder to extract semantic features, followed by a gripper-tip decoder that upsamples the feature maps using transposed convolution layers and outputs a two-channel heatmap corresponding to the left and right gripper tips. The network is trained by a Chamfer distance loss to balance localization accuracy and robustness under left–right symmetric ambiguities. Given predicted tip positions $\hat{\mathbf{p}_{tip}}$ and ground truth $\mathbf{p_{tip}}$, the Chamfer loss is formulated as:
\begin{equation} \begin{aligned} \mathcal{L}_{\text{chamfer}} =\; & \frac{1}{\left|\hat{\mathbf{p}}_{\text{tip}}\right|} \sum_i \min_j \left\|\hat{\mathbf{p}}_{\text{tip}}^{(i)} - \mathbf{p}_{\text{tip}}^{(j)}\right\|_2^2 \\ + & \frac{1}{\left|\mathbf{p}_{\text{tip}}\right|} \sum_j \min_i \left\|\mathbf{p}_{\text{tip}}^{(j)} - \hat{\mathbf{p}}_{\text{tip}}^{(i)}\right\|_2^2  \end{aligned} \end{equation} 
During inference, the detected tip locations provide an initial pose hypothesis for subsequent pose optimization.
\subsubsection{Pose Estimation and Tracking}
Estimating accurate articulated poses of surgical instruments from monocular video is challenging due to complex and large inter-frame motion. Inaccurate pose estimates not only degrade 2D-3D alignment but also severely affect subsequent appearance learning. To address this issue, we formulate instrument pose estimation as an optimization problem that directly aligns the rendered part-aware semantics map with observed segmentation through differentiable rendering. Specifically, given a candidate pose $q_i$ for the $i$-th frame, the semantics-embedded Gaussian $\mathcal{G}$ is transformed via forward kinematics and renders a part-aware semantic silhouette $\widehat{\mathbf{S}}_i$, which is compared against the observed instrument segmentation. Pose estimation is solved by minimizing the following objective:
\begin{equation}
\hat{\mathbf{q}} = \arg\min_{\mathbf{q}} \ \mathcal{L}_{pose}(\mathbf{q};\;\mathcal{G}),
\end{equation}
where the overall loss consists of a semantics alignment term and a tip matching constraint $\mathcal{L}_{tip}$:
\begin{equation}
\mathcal{L}_{pose} = \mathcal{L}_{sil} + \lambda_{tip} \mathcal{L}_{tip}
\end{equation}
The silhouette loss enforces global consistency between the rendered part-aware semantic silhouette $\widehat{\mathbf{S}}$ and the observed segmentation mask $\mathbf{S}$ with an $\ell_1$ distance: $\mathcal{L}_{{sil}} = \left\| \widehat{\mathbf{S}}_i - \mathbf{S}_i \right\|_{1}$.

While this term provides effective supervision for pose estimation, it alone cannot reliably align the small and left–right symmetric gripper tips. Let $\mathcal{T}_r$ and $\mathcal{T}_o$ denote the rendered and observed tip sets in the image plane. We define a symmetric nearest-neighbor loss
\begin{equation}
\mathcal{L}_{tip} =
\sum_{u \in \mathcal{T}_r} \rho\!\left(\min_{v \in \mathcal{T}_o}\|u-v\|_2\right)
+
\sum_{v \in \mathcal{T}_o} \rho\!\left(\min_{u \in \mathcal{T}_r}\|v-u\|_2\right),
\end{equation}
where $\rho(d)=\max(d-r,0)$ is a truncated penalty with tolerance radius $r$ to suppress unstable gradients caused by small localization noise. This symmetric formulation avoids explicit left–right assignment and improves robustness under symmetry ambiguity.

For pose tracking across frames, robust initialization is critical. The first frame is initialized via a small set of manually selected 2D-3D correspondences on the wrist region using a PnP solver. For subsequent frames, we employ a matching-based initialization strategy: features extracted from the wrist are matched between consecutive frames, lifted to 3D using depth rendered from the GS model, and used as 2D-3D correspondences to compute a coarse pose estimate via PnP. 

\subsection{Robust Texture Learning under Noisy Poses}
\label{sec:3.4}
\begin{figure}[t]
	\centering
	\includegraphics[width=\columnwidth]{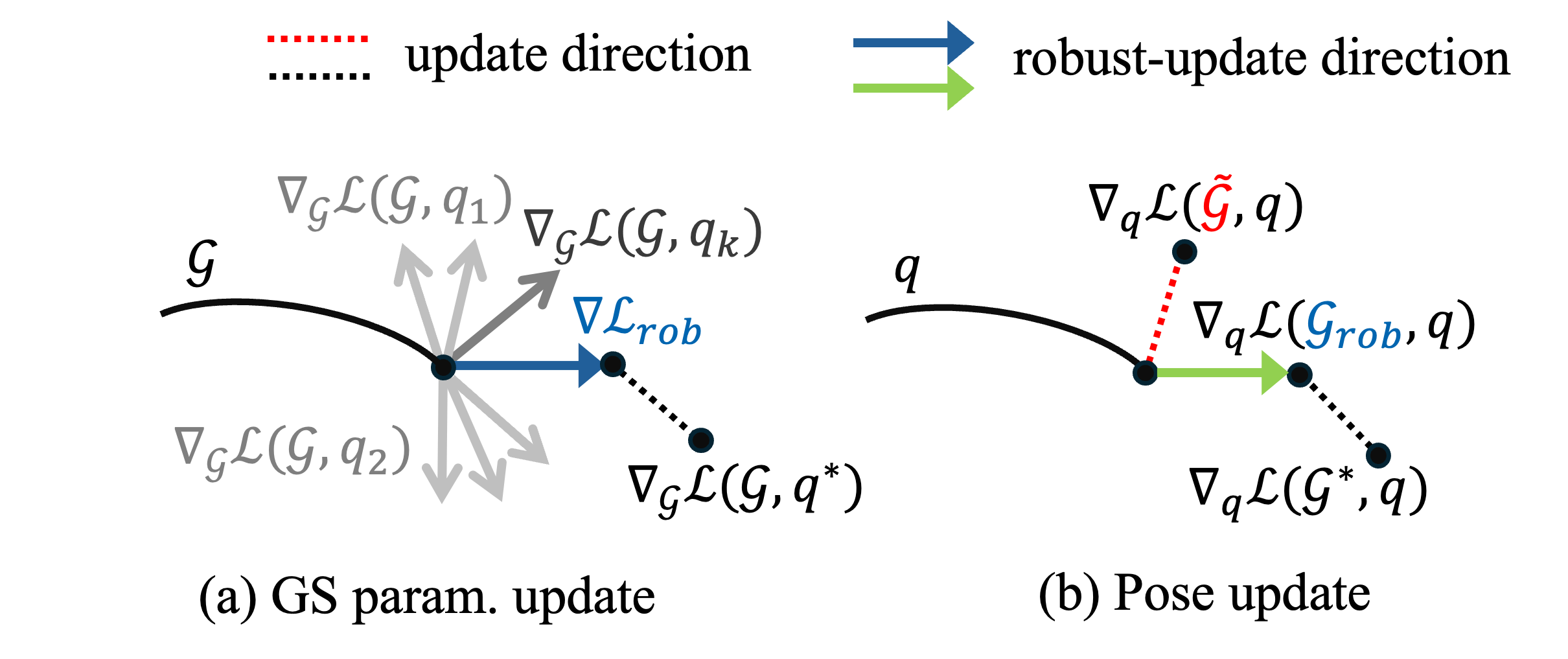}
	\vspace{-3mm}
	\caption{Left: Gaussian update under perturbed instrument poses. Right: The pose update is affected by the quality of Gaussian model. }

	\label{fig:robust}
	\vspace{-3mm}
\end{figure}
Estimated poses by SAPET inevitably contain minor noises due to imperfect segmentation. Directly using such imperfect poses to supervise the texture learning of our Gaussian model leads to suboptimal reconstruction quality. Driven by this issue, we propose a Robust Texture Learning (RTL) strategy that refines per-frame poses while explicitly mitigating the impact of noisy pose during texture learning. Rather than jointly optimizing poses and GS parameters, RTL contains two alternately-performed steps: pose refinement and robust Gaussian appearance optimization. 
\subsubsection{Warm-up Training}
We first perform 800 iterations of Gaussian parameter optimization to learn the coarse appearance of the surgical instrument. During this stage, the estimated poses obtained in Sec.~\ref{sec:3.3} are directly used without further refinement. The appearance learning is supervised by an $\ell_1$ photometric loss that minimizes the discrepancy between the rendered images and the corresponding ground-truth frames. To fully exploit the representation capability of GS, all Gaussian parameters are set to be trainable during this phase. In addition, a mask-based loss is introduced to preserve geometric fidelity. The training objective $\mathcal{L}_{tex}$ is formulated as:
\begin{equation}
\mathcal{L}_{tex} = \mathcal{L}_{\text{pho}} + \lambda_{\text{mask}}\mathcal{L}_{\text{mask}}
\end{equation}
\begin{equation}
\mathcal{L}_{\text{pho}} = \left\|  \widehat{\mathbf{I}}_i - \mathbf{I}_i \right\|_{1}, \quad
\mathcal{L}_{\text{mask}} = \left\| \widehat{\mathbf{M}}_i - \mathbf{M}_i \right\|_{1}
\end{equation}

\subsubsection{Pose Refinement}
After warm-up training, we refine the poses for each frame $I_i$ using the current Gaussian model before updating the appearance. Our core insight is that appearance learned from noisy poses still provides informative visual cues for further pose estimation. During pose refinement, we freeze the GS attributes $\mathcal{G}$ and refine the instrument pose $q$ over $K$ optimization steps:
\begin{equation}
    q_{i,k} \gets q_{i,k-1} - \eta \nabla_q \mathcal{L}_{tex}(\mathcal{G}, q_{i,k-1}),
    \label{eq:pose_ref}
\end{equation}
where $\nabla_q\mathcal{L}_{tex}$ denotes the gradient w.r.t. per-frame instrument pose, $q_{i,K}$ denotes the refined pose for the $i$-th frame, $K$ is set to 150 in our experiment. After the $K$-step pose refinement, the final pose $q_{i,K}$ approximates the optimal estimate under the current $\mathcal{G}$ for each frame $I_i$. 
\subsubsection{Robust Gaussian Appearance Optimization}
After pose refinement, a pose trajectory $\{q_{i,0},q_{i,1},\dots,q_{i,K}\}$ is formed for each frame $I_i$ while solving Eq.~\eqref{eq:pose_ref}. As illustrated in Fig.~\ref{fig:robust}, along this pose trajectory, we calculate the gradients w.r.t. the Gaussian parameters, forming a corresponding gradient sequence
$\{\nabla_{\mathcal{G}}\mathcal{L}_{tex}(\mathcal{G}, q_{i,0}), \nabla_{\mathcal{G}}\mathcal{L}_{tex}(\mathcal{G}, q_{i,1}), \dots, \nabla_{\mathcal{G}}\mathcal{L}_{tex}(\mathcal{G}, q_{i,K})\}$.
Each gradient is conditioned on a slightly different pose and therefore encodes distinct pose-dependent biases caused by imperfect alignment.

Rather than updating $\mathcal{G}$ using a single estimated pose $q_{i,K}$, we update it using gradients aggregated along the entire pose trajectory. Specifically, we define
\begin{equation}
\nabla \mathcal{L}_{\text{rob}}
= \mathbb{E}_{q \sim p_i(q)}\!\left[\nabla_{\mathcal{G}}\mathcal{L}_{\text{tex}}(\mathcal{G}, q)\right],
\end{equation}
where $\nabla \mathcal{L}_{\text{rob}}$ integrates gradients from multiple slightly perturbed poses, $p_i(q)$ is induced by the pose refinement trajectory $\{q_{i,k}\}_{k=0}^{K}$ and represents a local pose neighborhood. In practice, we approximate this $\nabla \mathcal{L}_{\text{rob}}$ by $\sum_{k=0}^{K} \nabla_{\mathcal{G}}\mathcal{L}_{\text{tex}}(\mathcal{G}, q_{i,k})$. Of note, we omit the scaling factor as it can be absorbed into the learning rate.
This aggregation reduces pose-induced gradient bias by stabilizing gradient direction under local pose neighborhood. This results in smoother optimization under pose uncertainty, providing a more reliable update direction in the Gaussian parameter space. We then update the Gaussian parameters using the aggregated gradient: $\mathcal{G'} = \mathcal{G} - \eta_{g} \, \nabla\mathcal{L}_{rob}$, where $\eta_{g}$ denotes the learning rate for Gaussian attributes, $\mathcal{G'}$ means the updated Gaussian model that is robust to small pose noises.
\subsubsection{Alternate Training}
The RTL optimization runs for 5K iterations per video during texture learning. In each iteration, we first perform $K$-step pose refinement while freezing Gaussian attributes $\mathcal{G}$ for a sampled image. Then, the Gaussian attributes $\mathcal{G}$ are updated using the trajectory-aggregated gradient. This alternating scheme decouples pose correction from appearance optimization. As training progresses, the progressively improved appearance of $\mathcal{G}$ provides more reliable photometric supervision, which in turn leads to increasingly accurate pose refinement.


\section{Experiments}
\begin{table*}[t]
\centering
\caption{\textbf{Quantitative comparison on intraoperative and in-house datasets. Intraoperative Dataset represents sequences from both EndoVis and SAR-RARP dataset. RMSE and RMSE-M represent depth error within entire image space and instrument mask, respectively.}}
\setlength{\tabcolsep}{2.2pt}
\renewcommand{\arraystretch}{1.}
\resizebox{0.8\textwidth}{!}{
\begin{tabular}{lccccc|ccccc}
\hline
\multirow{2}{*}{Method} &
\multicolumn{5}{c|}{Intraoperative} &
\multicolumn{5}{c}{In-house} \\
\cline{2-11}
 & PSNR$\uparrow$ & SSIM$\uparrow$ & LPIPS$\downarrow$ & RMSE$\downarrow$ & RMSE-M$\downarrow$
 & PSNR$\uparrow$ & SSIM$\uparrow$ & LPIPS$\downarrow$ & RMSE$\downarrow$ & RMSE-M$\downarrow$ \\
\hline

WIM            & 22.34 & 92.75 & 0.170 & 25.7 & 16.5 & 23.44 & 92.15 & 0.098 & 19.8 & 16.1 \\
SC-GS          & 20.42 & 89.48 & 0.146 & 19.8 & 11.3 & 24.41 & 93.28 & 0.070 & 17.4 & 15.6 \\ 
Shape-of-motion& 22.35 & 93.46 & 0.088 & 13.2 & \textbf{8.1} & 24.88 & 94.34 & 0.067 & 20.5 & \textbf{6.6} \\ \hdashline
TTGS           & 22.98 & 93.29 & 0.093 & 18.8 & 16.1 & 25.45 & 94.46 & 0.078 & 23.5 & 14.0 \\
EndoGaussian   & 22.39 & 93.64 & 0.089 & 27.9 & 19.8 & 25.26 & 93.85 & 0.075 & 25.2 & 14.8 \\
SurgicalGS      & 22.78 & 92.73 & 0.179 & 27.6 & 15.4 & 23.95 & 91.5 & 0.147 & 23.3 & 27.7 \\
Deform3DGS     & 24.03 & 93.33 & 0.101 & 24.6 & 11.8 & 25.48 & 93.37 & 0.055 & 27.6 & 22.8 \\ \hline
w/o GP         & 21.56 & 89.55 & 0.132 & 13.1 & 23.2 & 23.13 & 91.10 & 0.087 & 18.3 & 21.5 \\
w/o SAPET & 23.18 & 94.17 & 0.081 & 4.5& 9.4 & 25.38 & 93.89 & 0.077 & 5.9 & 10.9 \\
w/o RTL   & 25.38 & 95.70 & 0.046 & 4.4 & 9.4 & 28.25 & 95.33 & 0.048 & 5.7 & 9.4 \\

\textbf{Ours}  & \textbf{27.58} & \textbf{96.41} & \textbf{0.041} & \textbf{4.5} & 9.2
                & \textbf{29.82} & \textbf{96.27} & \textbf{0.039} & \textbf{5.6} & 9.4 \\
\hline
\end{tabular}}
\label{tab:sota}
\end{table*}
\begin{figure*}[h]
    \centering
    \includegraphics[width=0.8\textwidth]{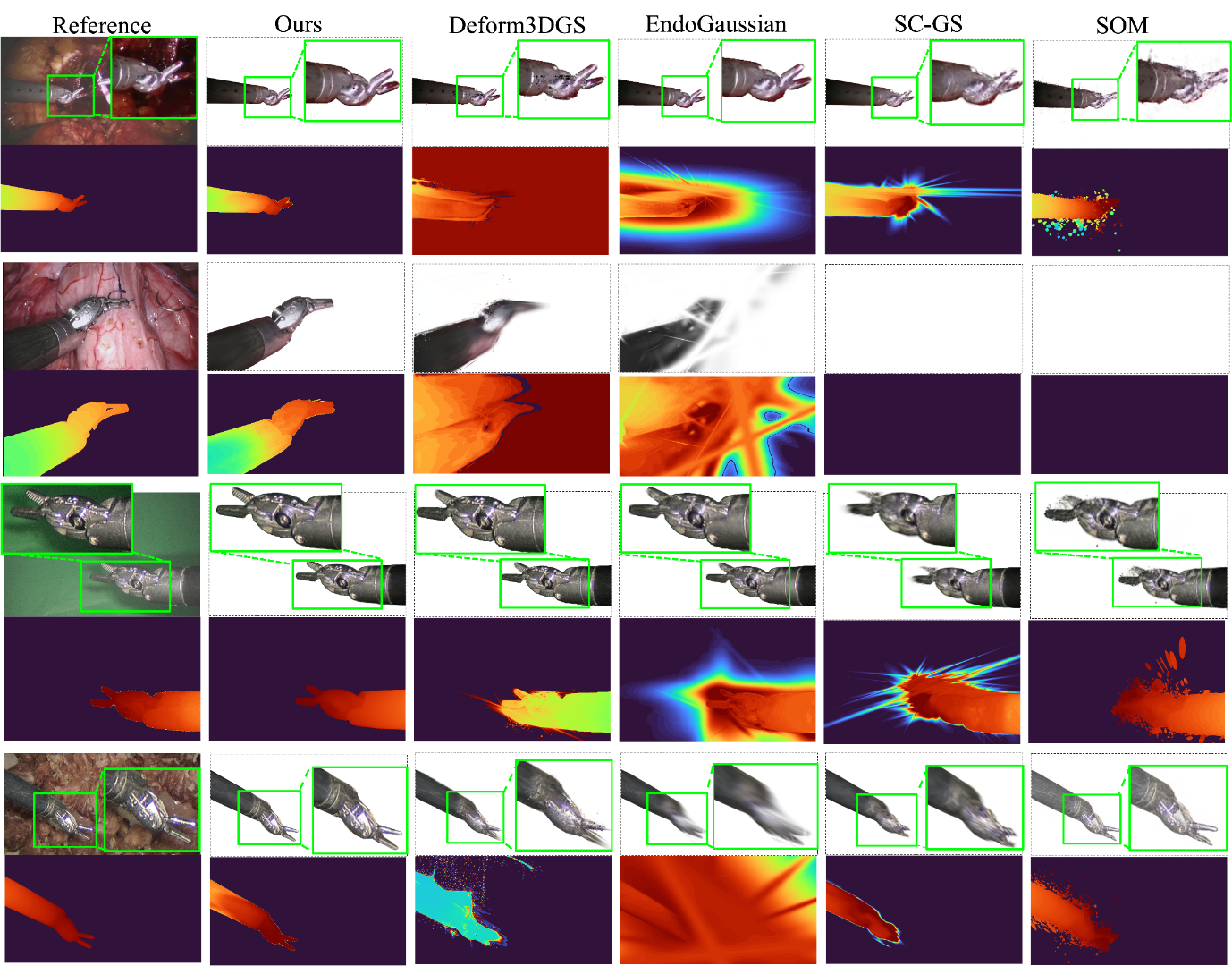}
    \caption{Visualization of the texture learning results on testing sets in comparison with SOTA reconstruction methods. We visualize the RGB and depth rendering. For SAR-RARP sequence (the first row), we use the depth rendering from CAD mesh with the estimated pose as the reference depth. For the other dataset, stereo depth estimations serve as the reference depth.   }
    \label{fig:texture}
\end{figure*}
\subsection{Datasets and Implementation}

\noindent \textbf{EndoVis datasets.}
We evaluate on EndoVis 2017~\cite{allan20192017} and EndoVis 2018~\cite{allan20202018}, which provide stereo endoscopic images with calibrated cameras and part-level annotations at 1-2 FPS. For each dataset, one trajectory of 100 frames ($512\times640$) is used for evaluation.

\noindent \textbf{SAR-RARP dataset.}
We further evaluate on the SAR-RARP dataset~\cite{psychogyios2023sar}, which contains monocular videos with realistic instrument-tissue interaction and complex intraoperative scenes. Two videos (four LND trajectories, 118 frames each) are sampled and resized from $1080\times1920$ to $360\times640$. As camera intrinsics are unavailable, a constant focal length of 571 and centered principal point are assumed following~\cite{allan20192017}. Part-level masks are generated using SAM2~\cite{ravi2024sam}.

\noindent \textbf{In-House dataset.}
We collect four stereo surgical videos with \textit{ex vivo} tissue and green-screen settings. Two green-screen and four tissue-background trajectories are extracted, with stereo images and five annotated 2D keypoints for pose evaluation. Five keypoints are defined on the instrument surface and are annotated using Indocyanine green (ICG) dyes visible only under fluorescent light. Data collection follows the SurgPose protocol~\cite{wu2025surgpose}, and part masks are obtained using SAM2~\cite{ravi2024sam}.

\noindent \textbf{Implementation Details.}
Geometry pretraining, pose estimation, and texture learning are performed sequentially. Each part is pretrained for 10K iterations, followed by per-frame pose optimization (up to 2K iterations with early stopping); only the first frame requires manual initialization. Loss weights are set to $\lambda_{mask}=0.1$ and $\lambda_{tip}=10^{-4}$. Texture learning is performed for 5K iterations per video and each video sequence is divided into seen and unseen poses with a 7:1 ratio for training and testing. During inference, per-frame poses of the testing set are first estimated by SAPET and then used as user-defined inputs to repose the textured instrument Gaussian. Next, quantitative evaluation is conducted by rendering all frames in the testing set with the estimated poses. Visual fidelity is assessed using PSNR, SSIM, and LPIPS. To evaluate geometric accuracy, we compute depth RMSE over the entire image (capturing errors from floating artifacts outside the instrument mask) and RMSE-M within the instrument mask to assess in-region depth errors. All experiments are implemented in PyTorch and run on a single NVIDIA RTX A5000 GPU.
\subsection{Comparison with State-of-the-art Methods}
\label{sec:sota}
We compare the proposed method with several SOTA approaches, including natural-scene reconstruction methods including WIM~\cite{noguchi2022watch}, SC-GS~\cite{huang2024sc}, and Shape-of-Motion (SOM)~\cite{wang2025shape}, as well as dynamic scene reconstruction baselines in surgical domain: SurgicalGS~\cite{chen2025surgicalgs}, TTGS~\cite{xu2025t}, EndoGaussian~\cite{liu2024endogaussian}, and Deform3DGS~\cite{yang2024deform3dgs}. Among nature-scene reconstruction methods, WIM is a NeuS~\cite{wang2021neus}-based framework designed for multi-view articulated robot reconstruction. SC-GS leverages sparse control points to drive neighboring GS points to model dynamic scenes. SOM uses point tracking to calculate dense point trajectories across video frames, facilitating the dynamic reconstruction under large motions. The remaining methods focus on dynamic surgical scenes with high visual fidelity but do not produce fully controllable articulated assets; nevertheless, they provide strong baselines for reconstruction quality evaluation.

Tables~\ref{tab:sota} reports the quantitative results on the intraoperative (EndoVis17\&18 and SAR-RARP) and in-house datasets, respectively. Compared with SOTA methods for natural dynamic scenes and surgical scenarios, our approach achieves consistently higher photometric accuracy, demonstrating superior appearance reconstruction under complex instrument motions. Although SOM and TTGS introduce point tracking to better capture instrument motion and improve over earlier methods, they still remain inferior to our kinematics-aware framework. As for geometric fidelity, existing methods struggle to recover accurate structures from RGB-D surgical videos, leading fragmented or artifact-prone geometry as in Fig.~\ref{fig:texture}. Notably, SOM achieves the lowest RMSE-M, indicating its ability to capture fine-grained depth variations, while our method provides better RMSE, representing more physically plausible geometry overall. In Fig.~\ref{fig:texture}, under large inter-frame motions in EndoVis videos, general-domain methods such as SOM and SC-GS fail to produce valid reconstructions, resulting in empty outputs (second row). Although existing methods achieve visually comparable results in scenes with smoother motion (first, third, and fourth rows), they suffer from significant geometric distortions, limiting the usability for physical simulation.
\subsection{Ablation Study on Key Components}
\begin{figure}[t]
	\centering
	\includegraphics[width=\columnwidth]{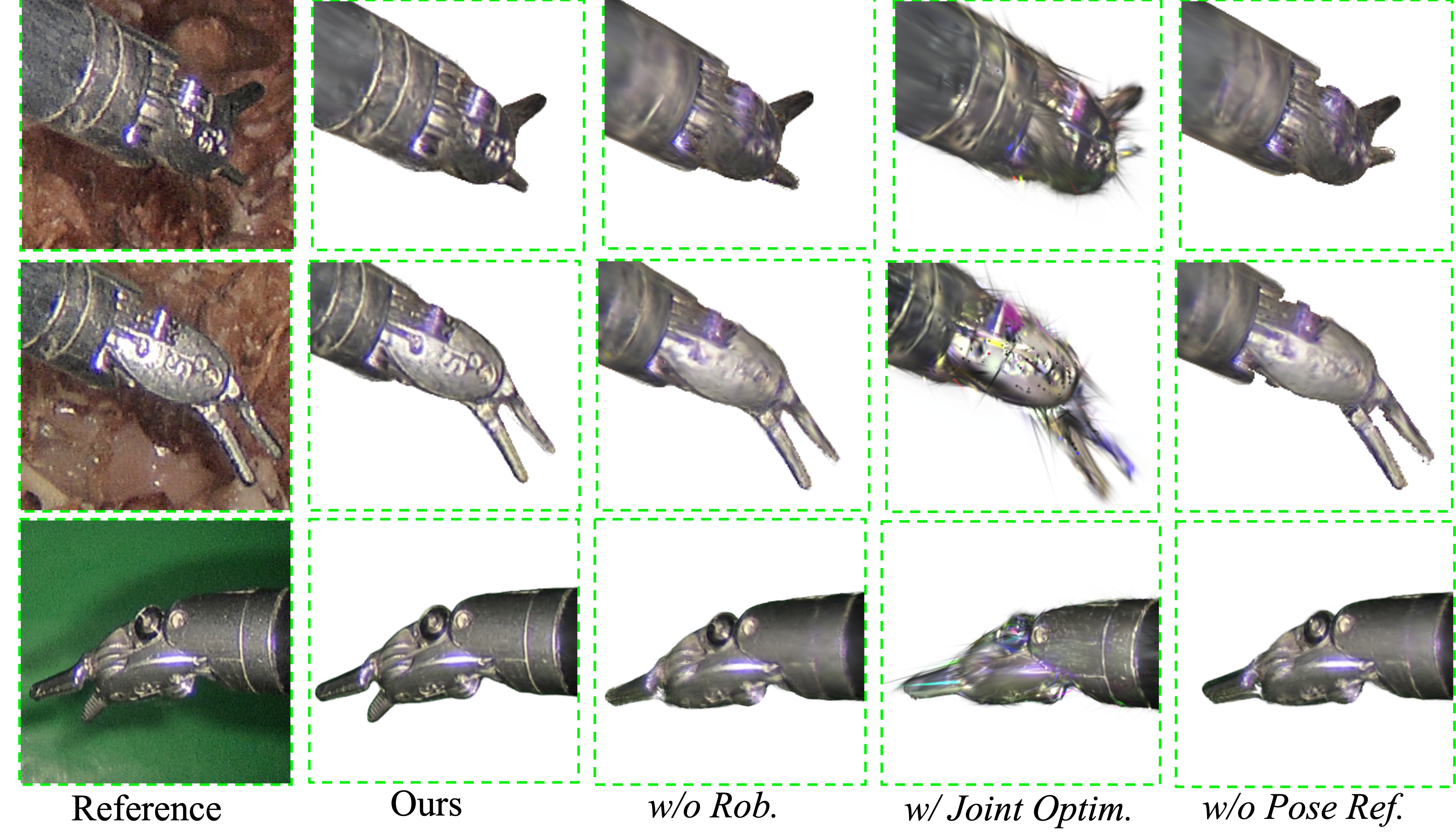}
	\vspace{-3mm}
	\caption{Visualization of the variant methods on texture learning.}

	\label{fig:ablation}
	\vspace{-3mm}
\end{figure}
To analyze the contribution of each component in our framework, we conduct ablation studies by disabling geometry pretraining (\textit{w/o GP}), where GS primitives for each part are initialized from randomly sampled points; replacing our semantic-aware pose estimation and tracking strategy (including Gripper-Tip Net) ) by a naive render-and-compare pose estimation module that solely uses the silhouette mask to optimize instrument pose (\textit{w/o SAPET}); removing the proposed robust texture learning strategy (\textit{w/o RTL}) and instead, using the estimated poses with noises to perform texture learning as in \cite{yang2025instrument}.

Quantitative results on both datasets (Table~\ref{tab:sota}) demonstrate that each component is essential for stable and high-fidelity reconstruction. Removing geometry pretraining causes a substantial drop in both photometric and depth accuracy, indicating that accurate geometric initialization is critical for preserving instrument structure. Disabling robust texture learning leads to increased reconstruction noise and degraded depth accuracy, confirming its effectiveness in alleviating the adverse effects of pose perturbations. Replacing our SAPET with a naive render-and-compare pipeline also significantly degrades the reconstruction performance, indicating that decreased pose estimation accuracy strongly compromises the texture learning and validates the effectiveness of our well-designed SAPET pipeline. Finally, the results of \textit{w/o RTL} demonstrate that our RTL can further enhance visual fidelity on both intraoperative and in-house dataset.

\begin{table}[t]
\centering
\caption{\textbf{Analysis of texture learning on In-House Dataset.} PSNR, SSIM, and LPIPS measure the photometric accuracy while RMSE measures depth accuracy.}
\setlength{\tabcolsep}{3pt}   
\renewcommand{\arraystretch}{1.08} 
\resizebox{0.8\linewidth}{!}{%
\begin{tabular}{ccccc}
\hline
Method        & \makecell{PSNR$\uparrow$\\} &  \makecell{SSIM$\uparrow$\\} &  \makecell{LPIPS$\downarrow$\\} &  \makecell{RMSE$\downarrow$} \\ \hline
\textit{w/o pose ref.}        & 29.24 & 95.69 & 0.051 & 5.6  \\
\textit{w/o Rob}    & 28.64 & 95.51 & 0.047 & 5.6\\
\textit{w/ joint Optim.}  & 22.48 & 91.17 & 0.068 & 5.9  \\
\textit{RTL}   & \textbf{29.82} & \textbf{96.27} & \textbf{0.039} & \textbf{5.6}  \\ \hline
\end{tabular}%
}
\label{tab:texture}
\end{table}
\subsection{Detailed Analysis}
\subsubsection{Analysis on Robust Texture Learning}
We then investigate the effectiveness of RTL design on the in-house dataset by removing pose refinement step (\textit{w/o pose ref.}) and robust appearance optimization (\textit{w/o Rob.}), respectively. We also compare our alternating training scheme with a commonly used joint optimization strategy (\textit{w/ joint optim.}) that simultaneously updates pose and Gaussian parameters. Quantitative and qualitative results are shown in Table.~\ref{tab:texture} and Fig.~\ref{fig:ablation}, respectively. Excluding pose refinement and robust appearance optimization degrade reconstruction quality, verifying the effectiveness of both submodules in our RTL. Joint optimization tends to overfit training viewpoints by distorting the underlying instrument geometry, resulting in severe degradation under unseen poses. As illustrated in Fig.~\ref{fig:ablation}, pose refinement corrects misaligned pose (gripper state) while training proceeds; joint optimization produces obvious geometric distortions and visual artifacts. In addition, reconstructions from both \textit{w/o Rob} and \textit{w/o pose ref.} exhibit noticeable blurring on the wrist surface, reflecting suboptimal texture learning under noisy pose estimates. 
\subsubsection{Analysis on Pose Estimation Accuracy}
\begin{table}[t]
\centering
\caption{\textbf{Analysis on pose estimation and tracking performance.}
Comparison of model variants on EndoVis17\&18 and the In-House dataset.}
\vspace{-1mm}
\renewcommand{\arraystretch}{1.05} 
\resizebox{0.9\linewidth}{!}{%
\begin{tabular}{lccccc}
\toprule
\multirow{2}{*}{\textbf{Method}} &
\multicolumn{3}{c}{\textbf{EndoVis17\&18 (Dice $\uparrow$)}} &
\multicolumn{2}{c}{\textbf{In-House (Keypoint)}} \\
\cmidrule(lr){2-4}\cmidrule(l){5-6}
 & Shaft & Wrist & Gripper & \multicolumn{1}{c}{\makecell{RMSE$\downarrow$\\(pix)}} & \makecell{RMSE$\downarrow$\\(mm)} \\
\midrule
w/o Matching       & 93.73 & 83.12 & 66.03 & 9.79 & 8.64 \\
w/o Reg & 95.52 & 78.40 & 63.39 & 6.36 & 5.45 \\
w/ SVD         & \textbf{96.83} & 86.65 & 73.32 & 5.85 & 5.12 \\
w/o Gripper-Tip Net        & 94.63 & 83.94 & 63.17 & 10.11 & 7.58 \\
SAPET       & 95.92 & 87.54 & 75.27 & 4.75 & \textbf{3.69} \\
SAPET+Pose Ref.        & 96.21 & \textbf{87.68} & \textbf{75.94} & \textbf{3.91} & 3.73 \\

\bottomrule
\end{tabular}%
} 
\vspace{-2mm}
\label{tab:pose_ablation}
\end{table}
\begin{figure*}[t]
    \centering
    \includegraphics[width=0.9\textwidth]{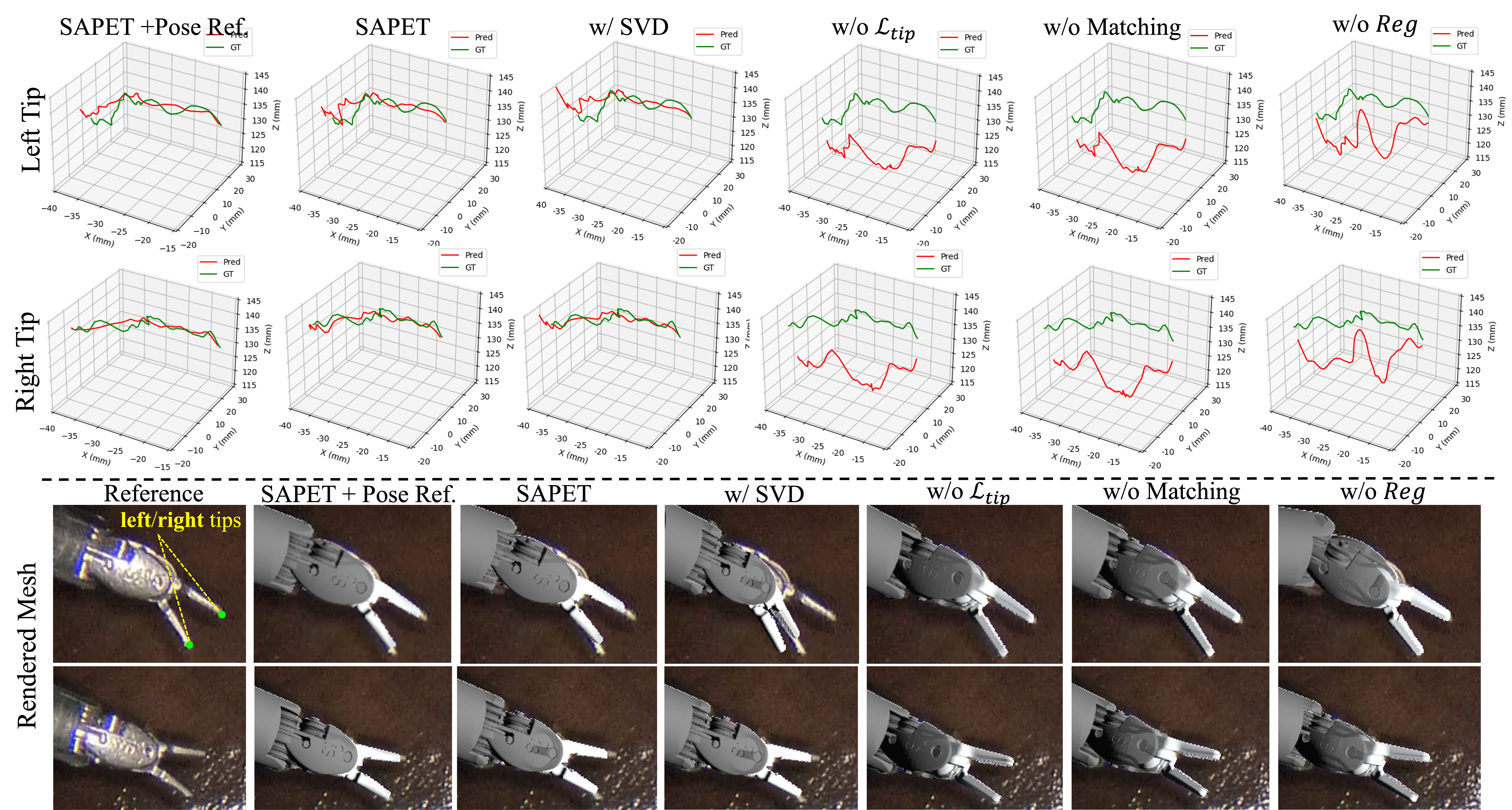}
    \caption{Visualization of the pose estimation results on in-house dataset. We visualize the predicted and GT trajectories of the left and right gripper tip points across a 100-frame video clip. The predicted and GT trajectories are colored in green and red, respectively. With estimated poses, we render the mesh to 2D original images to visualize the reprojection errors at frame 2 and 58, respectively.   }
    \label{fig:pose_ablation}
\end{figure*} 
We further investigate the pose estimation accuracy of our proposed SAPET and pose refinement. As ground-truth kinematics are unavailable in existing benchmarks, pose accuracy is evaluated using two complementary metrics: (1) Dice between rendered part-aware maps and ground-truth segmentation on EndoVis17\&18, and (2) 2D/3D keypoint RMSE on our in-house dataset with five annotated landmarks. Stereo depth~\cite{cheng2025monster} is used to lift 2D keypoints to 3D. Quantitative and qualitative results are shown in Table~\ref{tab:pose_ablation} and Fig.~\ref{fig:pose_ablation}. 

The full pose estimation and tracking pipeline as in Sec.~\ref{sec:3.3} (\textit{SAPET}) achieves the best overall performance, indicating that the proposed submodules jointly provide accurate and stable pose estimation. Without Gripper-Tip Net (\textit{w/o Gripper-Tip Net}), the pose estimation accuracy degrades significantly. Also, replacing Gripper-Tip Net with  closed-form SVD-based tip estimation (\textit{w/ SVD}) leads to decreased Dice scores on regions with dexterous motions and keypoint localization accuracy.  Finally, incorporating the pose refinement in robust texture learning (\textit{SPET+Pose Ref.}) further improves Dice scores and 2D localization; the slight degradation observed in 3D keypoint RMSE is attributed to stereo depth measurement noise. This pose refinement is performed during the texture learning stage by optimizing poses using photometric loss with the learned Gaussian appearance. The results demonstrate that learned texture provides additional fine-grained visual constraints for pose optimization, validating the complementary role of appearance in enhancing pose estimation accuracy. 
\subsection{Instrument-Splatting++ Benefits Keypoint Detection}
\label{sec:keypoint}
With high controllability, we can evaluate textured Gaussian's effectiveness in augmenting downstream surgical instrument keypoint detection following~\cite{wu2025surgpose,ghanekar2025video} via unseen-pose data generation. 
\subsubsection{Experimental Setting}
Keypoint detection is to localize 5 pre-defined keypoints on the instrument surface from RGB images. We select a 100-frame video with \textit{ex vivo} tissue background from the in-house dataset as the real training set and a 1,000-frame test set sampled from SurgPose dataset\cite{wu2025surgpose} with different instruments, backgrounds, and motion patterns for evaluation. We will apply our method to estimate per-frame poses and reconstruct a textured instrument Gaussian from the 100-frame real training set. Next, this textured GS will be used to generate a synthetic dataset under unseen poses as data augmentation.
\subsubsection{Unseen-Pose Dataset Synthesis}
\begin{figure}[t]
\centering
    \includegraphics[width=\columnwidth]{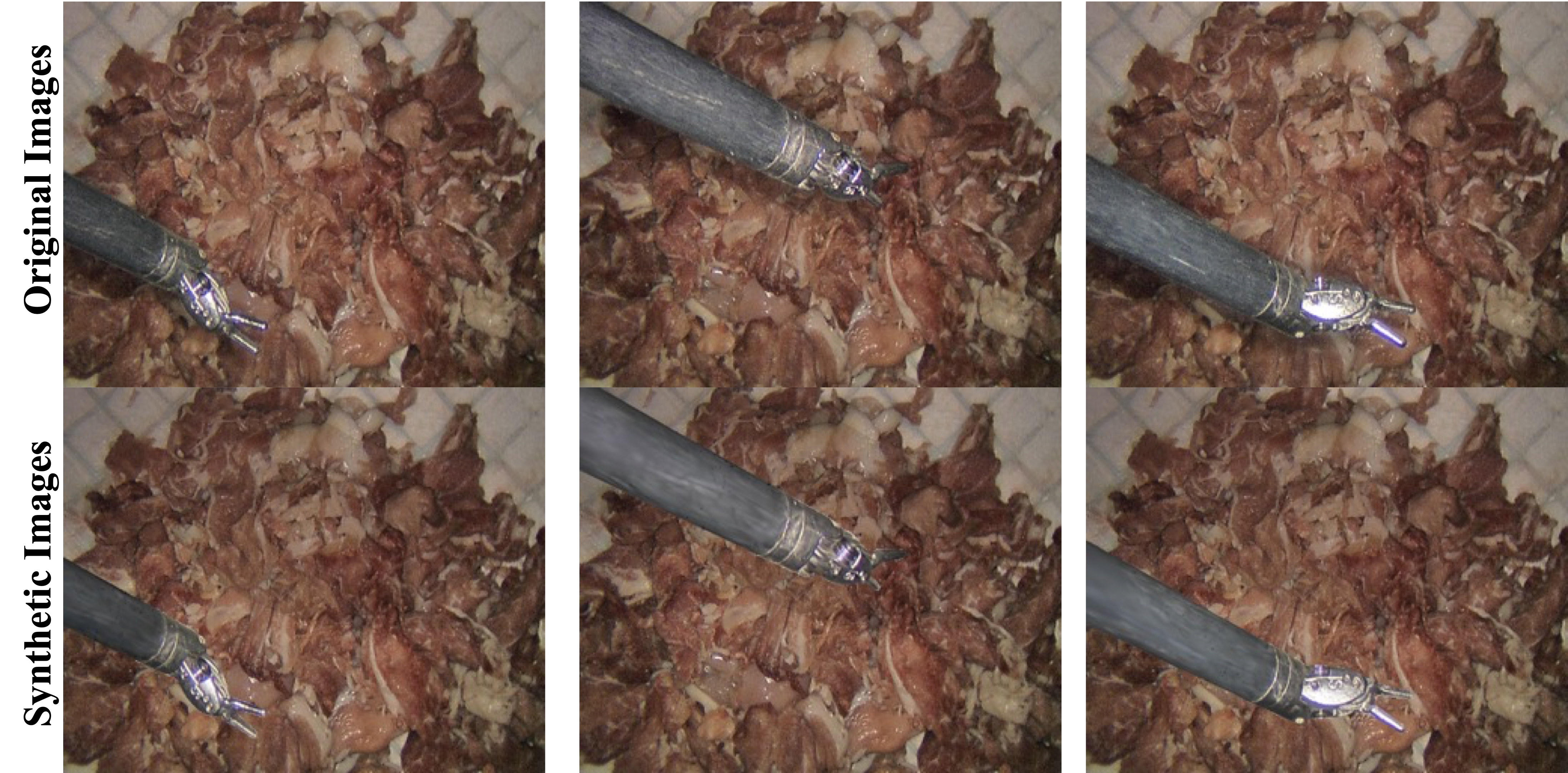}
    \caption{Visual comparison between synthetic and reference original images. Tissue and instrument GS are both learned from the real training set. Reference images are selected from another video with different instrument poses, which are estimated by SAPET and serve as user-defined input to repose the instrument GS.}
    \label{fig:combination}
\end{figure}
We generate synthetic training data by randomly perturbing the estimated poses to repose the reconstructed instrument GS. To avoid occlusion artifacts, the tissue background is independently reconstructed using Deform3DGS~\cite{yang2024deform3dgs} after masking out the instrument, and the final synthetic scenes are obtained by compositing the reposed instrument Gaussian with the learned tissue Gaussian. During data synthesis, we use the same background as the 100-frame real training set without introducing additional scene diversity. Key point GT for synthetic data is obtained by projecting predefined 3D keypoints onto the image plane through forward kinematics. In total, 10,000 synthetic images are generated. Visualization results in Fig.~\ref{fig:combination} show that the synthetic images closely match reference images from another video with different instrument trajectories. More results with unseen poses are in the supplementary video.
\subsubsection{Results}
\begin{table}[t]
\centering
\caption{\textbf{Downstream keypoint detection on surgical instruments.}
Accuracy$\uparrow$ measures detection accuracy; RMSE$\downarrow$ and PCK@pixel$\uparrow$ report pixel error.}
\setlength{\tabcolsep}{3pt}   
\renewcommand{\arraystretch}{1.08} 
\resizebox{0.8\linewidth}{!}{%
\begin{tabular}{ccccc}
\hline
Data        & \makecell{Accuracy$\uparrow$\\(\%)} &  \makecell{RMSE$\downarrow$\\(pix)} &  \makecell{PCK@2.5$\uparrow$\\(\%)} &  \makecell{PCK@5$\uparrow$\\(\%)} \\ \hline
\textit{Real}        & 97.4 & 6.71 & 65.71 & 86.15 \\
\textit{Syn}   & 95.6 & 6.55 & 52.63 & 77.54 \\
\textit{Real* + Syn} & 98.08 & 3.92 & 70.14 & 88.70 \\
\textit{Real + Syn}  & \textbf{98.16} & \textbf{3.23} & \textbf{73.57} & \textbf{90.62} \\ \hline
\end{tabular}%
}
\label{tab:downstream_kp}
\end{table}
\begin{figure}[t]
    \centering
    \includegraphics[width=1.\columnwidth]{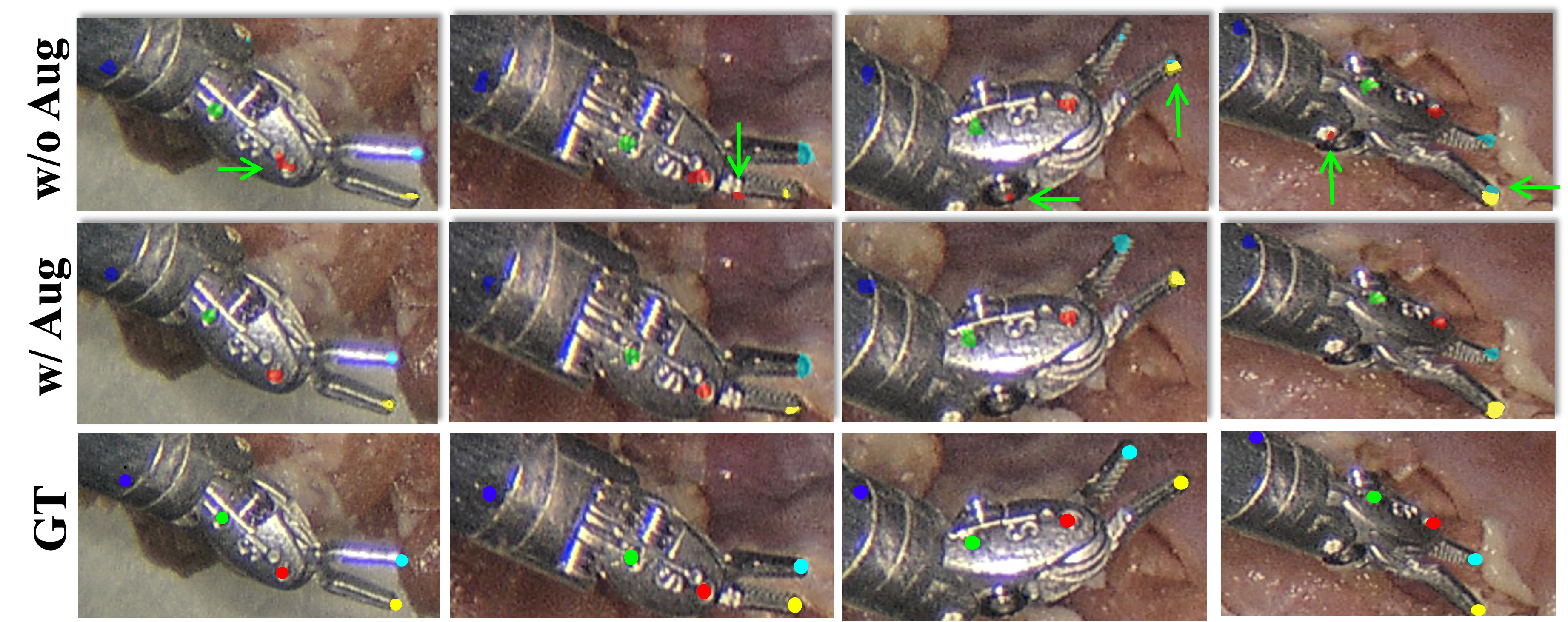}
    \caption{Qualitative visualization of downstream keypoint detection task on the testing set. We visualize estimated keypoints with/without unseen-pose data augmentation during training. }
    \label{fig:downstream_vis}
\end{figure}
For downstream keypoint detection, we train the model in \cite{ghanekar2025video} which predicts five keypoint heatmaps. As summarized in Table~\ref{tab:downstream_kp}, \textit{Real} denotes training on the 100-frame real dataset with ground-truth labels, \textit{Syn} refers to synthetic augmentation data generated by reposing instrument Gaussian, \textit{Real* + Syn} relies on pseudo labels projected from estimated kinematics on real dataset for fully unsupervised training, and \textit{Real + Syn} combines real data with known GT keypoints with synthetic augmentation. For each dataset combination, model is trained for sufficient number of epochs to ensure convergence. Performance is measured using detection Accuracy, RMSE for pixel localization error, and PCK@2.5/5 (percentage of correct keypoints within 2.5 and 5 pixels from the GT). While training solely on synthetic data underperforms real data, pseudo-label supervision already achieves reasonable accuracy without manual annotation, highlighting the potential of our method for fully unsupervised learning. As shown in Table~\ref{tab:downstream_kp} and Fig.~\ref{fig:downstream_vis}, combining real data with synthetic augmentation consistently improves localization accuracy, demonstrating the effectiveness of controllable instrument Gaussian for unseen-pose data augmentation.

\section{Discussion}
Instrument-Splatting++ is a controllable Real2Sim framework for reconstructing articulated surgical instruments from monocular videos, and it shows clear advantages over existing surgical reconstruction methods. In the comparison study (Sec.~\ref{sec:sota}), existing SOTA dynamic reconstruction methods remain limited when handling articulated surgical instruments. Even with dense point tracking, approaches such as SOM and TTGS struggle to model complex motions involving axial rotations. During such motions, large portions of the instrument surface continuously disappear and reappear due to self-occlusion and disocclusion, breaking temporal correspondence. Meanwhile, the large textureless metallic regions provide weak visual cues. These factors lead to unstable tracking and incorrect structural modeling. Although mask or depth supervision have been introduced, these methods still produce fragmented geometry and floating artifacts in rendered depth, as observed in Fig.~\ref{fig:texture}. In contrast, our method incorporates explicit CAD-based geometry and kinematic constraints, which regularize the solution space and ensure physically plausible structure while learning appearance.

Inaccurate poses introduce misalignment during texture learning, and thus compromise the reconstructed quality. Despite this, the textured Gaussian learned under these imperfect poses already captures meaningful visual information. This insight motivates the pose refinement stage in RTL, where photometric consistency between learned appearance and real images is used to progressively correct pose estimates. Meanwhile, pose refinement and texture learning form a mutually reinforcing process: improved appearance enables more reliable pose optimization, while refined poses provide more accurate alignment for subsequent appearance learning. The results in Table~\ref{tab:texture} and Table~\ref{tab:pose_ablation} indicate that the photometric pose refinement improves both pose estimation accuracy and reconstruction quality. 

Beyond enable high-fidelity and controllable Real2Sim reconstruction, our framework also provides a practical solution for scalable pose labeling from monocular surgical videos. By combining SAPET with photometric pose refinement, Instrument-Splatting++ can recover per-frame articulated tool poses without robot kinematics or external tracking, effectively functioning as an automatic pose annotator. The resulting video–pose pairs enable weakly or unsupervised learning for downstream 3D perception tasks, similar to keypoint localization in Sec.~\ref{sec:keypoint}, where pose labels are often unavailable. In addition, this capability offers a practical data source for robot imitation learning in surgical autonomy, reducing the reliance on accurate kinematics logs. Besides, our method achieves rendering at 102 FPS, enabling real-time interactive simulation and flexible manipulation of the digital twin.
\section{Conclusion}
We presented \textit{Instrument-Splatting++}, a Real2Sim framework that reconstructs surgical instruments from monocular endoscopic videos as fully controllable 3D Gaussian digital twins. By integrating CAD-based geometry pretraining with explicit kinematic constraints, the reconstructed instrument maintains physically plausible articulation and can be reposed to user-defined configurations. Built on this representation, we introduced SAPET for semantics-aware pose estimation and tracking, together with Robust Texture Learning (RTL) to enable stable appearance optimization under inevitable pose noise. Experiments on public and in-house datasets demonstrate superior photometric quality and improved geometric consistency, while downstream results validate the effectiveness of the controllable instrument Gaussian for unseen-pose data augmentation. A current limitation is that the tissue digital twin is not yet controllable, due to the challenges of modeling soft-tissue deformation and instrument–tissue interactions under occlusions and specularities. Future work will focus on interaction-aware, controllable tissue modeling and tighter instrument–tissue coupling, enabling more comprehensive and physically grounded surgical Real2Sim environments.

\bibliographystyle{IEEEtran}
\small\bibliography{refs}

\end{document}